\documentclass[10pt,twocolumn,letterpaper]{article}
\PassOptionsToPackage{dvipsnames,table}{xcolor}
\usepackage{amsthm}
\usepackage{cvpr}              

\definecolor{cvprblue}{rgb}{0.21,0.49,0.74}
\usepackage[pagebackref,breaklinks,colorlinks,citecolor=cvprblue]{hyperref}

\def\paperID{10568} 
\def\confName{CVPR}
\def\confYear{2025}

\title{\textit{Coeff-Tuning}: A Graph Filter Subspace View for Tuning Attention-Based Large Models}

\author{Zichen Miao$^\dagger$, Wei Chen$^\dagger$, Qiang Qiu \\
Purdue University, IN, USA\\
{\tt\small \{miaoz, chen2732, qqiu\}@purdue.edu}
}

\begin{document}

\newtheorem{theorem}{Theorem}[section]
\newtheorem{proposition}[theorem]{Proposition}
\newtheorem{lemma}[theorem]{Lemma}
\newtheorem{corollary}[theorem]{Corollary}
\newtheorem{definition}[theorem]{Definition}
\newtheorem{assumption}[theorem]{Assumption}
\newtheorem{remark}[theorem]{Remark}
\newcommand{\method}{\textit{Coeff-Tuning}}

\newcommand{\Wq}{\mathbf{W}_q}
\newcommand{\Wk}{\mathbf{W}_k}
\newcommand{\Wv}{\mathbf{W}_v}
\newcommand{\Wo}{\mathbf{W}_o}
\newcommand{\Wvo}{\mathbf{W}_{vo}}
\newcommand{\Q}{\mathbf{Q}}
\newcommand{\K}{\mathbf{K}}
\newcommand{\V}{\mathbf{V}}
\newcommand{\W}{\mathbf{W}}
\newcommand{\attn}{\mathbf{A}}
\newcommand{\inp}{\mathbf{X}}
\newcommand{\inpv}{\mathbf{x}}
\newcommand{\out}{\mathbf{O}}
\newcommand{\feat}{\mathbf{Z}}
\newcommand{\filter}{\mathbf{F}}
\newcommand{\coeff}{\boldsymbol{\alpha}}

\maketitle
\renewcommand*{\thefootnote}{\fnsymbol{footnote}}
\footnotetext[2]{Equal contribution.}
\renewcommand{\thefootnote}{\arabic{footnote}}

\begin{abstract}
Transformer-based large pre-trained models have shown remarkable generalization ability, and various parameter-efficient fine-tuning (PEFT) methods have been proposed to customize these models on downstream tasks with minimal computational and memory budgets.
Previous PEFT methods are primarily designed from a tensor-decomposition perspective that tries to effectively tune the linear transformation by finding the smallest subset of parameters to train.
Our study adopts an orthogonal view by representing the attention operation as a graph convolution and formulating the multi-head attention maps as a convolutional filter subspace, with each attention map as a subspace element. 
In this paper, we propose to tune the large pre-trained transformers by learning a small set of combination coefficients that construct a more expressive filter subspace from the original multi-head attention maps. We show analytically and experimentally that the tuned filter subspace can effectively expand the feature space of the multi-head attention and further enhance the capacity of transformers. 
We further stabilize the fine-tuning with a residual parameterization of the tunable subspace coefficients, and enhance the generalization with a regularization design by directly applying dropout on the tunable coefficient during training.
The tunable coefficients take a tiny number of parameters and can be combined with previous PEFT methods in a plug-and-play manner. 
Extensive experiments show that our approach achieves superior performances than PEFT baselines with neglectable additional parameters\footnote{Code is available at: \url{https://github.com/ZichenMiao/Coeff_Tuning}}.
\end{abstract}

\section{Introduction}
\begin{figure}[t]
    \centering
    \includegraphics[width=0.98\linewidth]{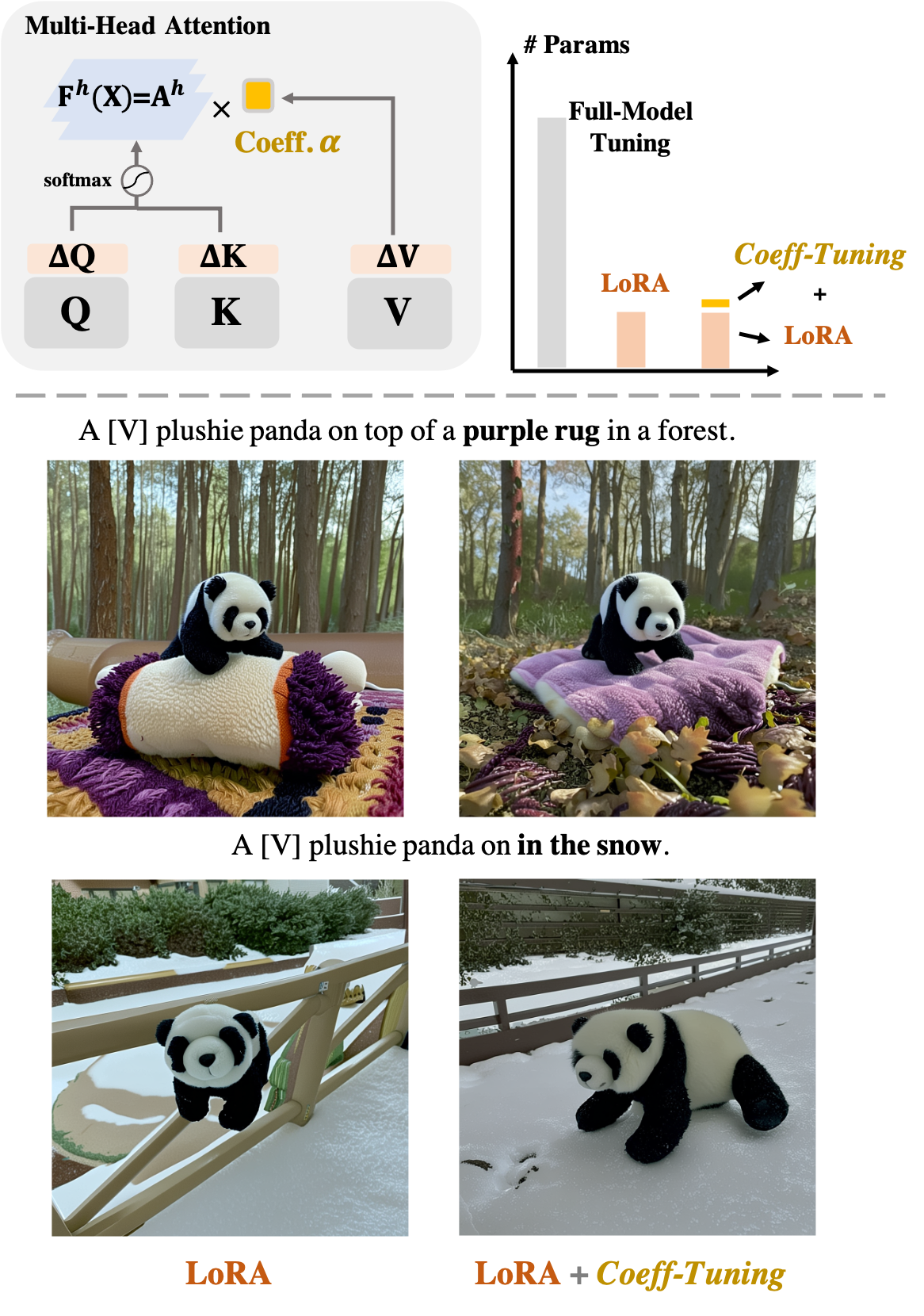}
    \caption{(Up)~Under the graph convolution view of the attention, the proposed~\method ~enhances the expressiveness by combining the attention maps with a tiny tunable subspace coefficient $\coeff$. It can be seamlessly integrated with other PEFT methods, e.g., LoRA~\cite{hu2021lora}, by introducing neglectable additional parameters. (Bottom) Combining \method~with LoRA achieves better results in the concept customization task with SDXL~\cite{podell2023sdxl}.}
    \label{fig:teaser}
    \vspace{-10pt}
\end{figure}

Large foundational models have demonstrated remarkable performance and generalization ability across a wide range of natural language processing and computer vision tasks~\cite{vit, vaswani2017attention, rombach2022high, sam}. The success of these models can be attributed to their ability to learn rich, transferable representations by leveraging the power of the transformer architecture~\cite{vaswani2017attention} and the extensive knowledge captured during pre-training on large-scale datasets~\cite{gpt, schuhmann2022laion}. However, fine-tuning these pre-trained transformers on downstream tasks with full-parameter updates can be challenging due to the high computational and memory costs. 


To address this, various parameter-efficient fine-tuning (PEFT) methods have been developed to adapt pre-trained models on downstream tasks while minimizing the number of trainable parameters~\cite{hu2021lora, liu2024dora, zaken2021bitfit, oft}. A representative family of these approaches focuses on tuning the linear transformation layers in transformers, where they apply a lightweight parameterization, e.g., low-rank parameterization~\cite{hu2021lora, liu2024dora}, to adapt the weight matrices. However, none of these methods are designed from a holistic view of the core operation in transformers, the multi-head attention layer~\cite{vaswani2017attention}, which plays a key role in the expressiveness of transformers~\cite{dong2021attention,shi2022revisiting}.

In this paper, we propose a new PEFT method tailored to enhance the multi-head attention's expressiveness by tuning a minimal set of parameters.
We first formulate the multi-head attention as a graph convolution with a filter subspace, with each graph filter as the attention map of each attention head.
We then tune the filter subspace by simply learning a small set of subspace coefficients that linearly combine the pre-trained attention maps. In this way, we adapt for each attention head a new task-adaptive attention map that better captures the token-wise correlation. 

We attribute the effectiveness of our approach to its ability to expand the feature space of the original multi-head attention layer. By applying the softmax function to the query-key correlation matrices, all elements in a row of the original attention map are non-negative and have a sum of one~\cite{vaswani2017attention}. This limits the expressiveness of the multi-head attention as the output token embeddings lie within the convex hull of the value embeddings. Our method constructs a new filter subspace with non-constrained coefficients as shown in Figure~\ref{fig:teaser}, allowing negative elements learned in the combined attention map to capture more diverse token-wise relationships. This also enables the output tokens to obtain representation outside the convex hull of the value embeddings, which further enhances the capacity of the multi-head attention. We provide analysis and an illustrative toy example to better demonstrate this point.



\begin{figure*}
    \centering
    \includegraphics[width=0.94\linewidth]{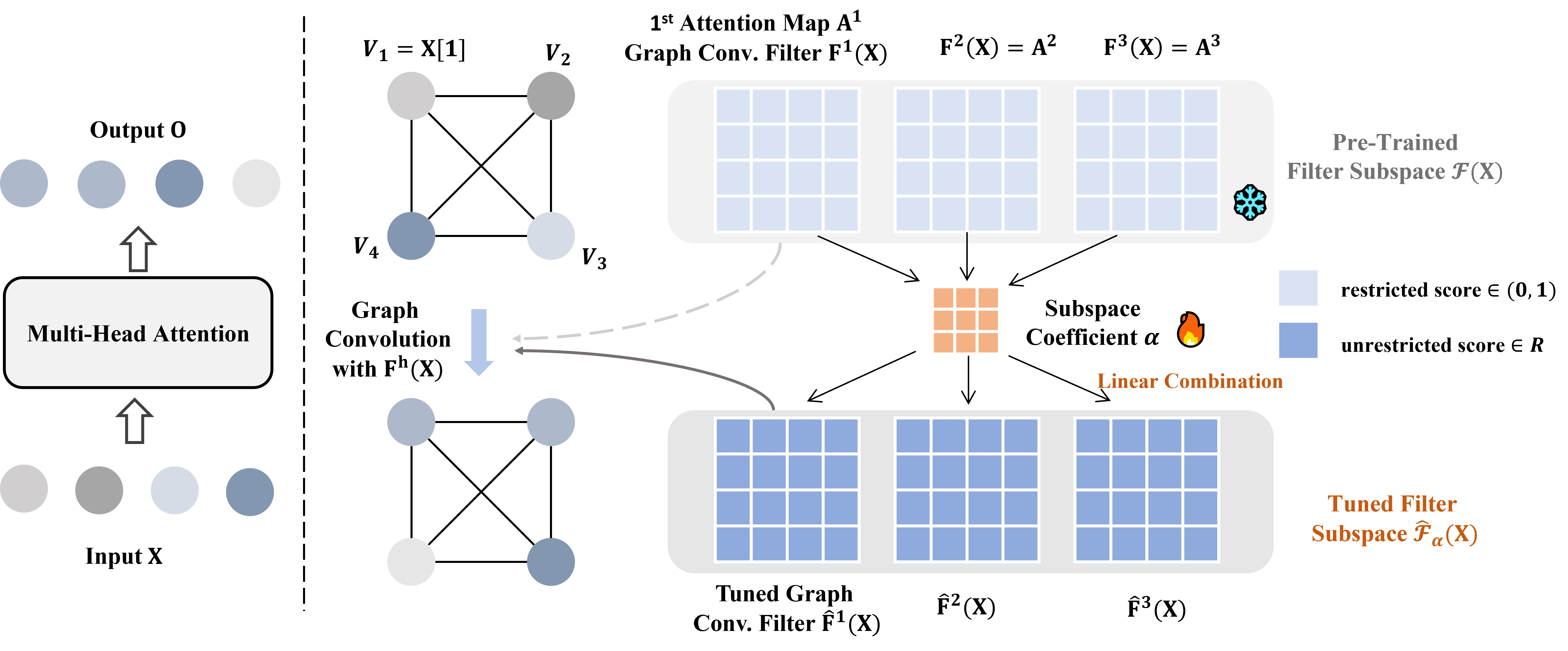}
    \caption{Illustration of the proposed \method. We first view each token $\mathbf{X}[i]$ as a node $V_i$ in an all-connected graph $\mathcal{G}=(V,E)$. The attention operation then turns into a graph convolution with each attention map $\mathbf{A}^h$ as a graph convolutional filter $\mathbf{F}^h(\mathbf{X})$. In this way, the attention maps in the multi-head attention layer form a filter subspace $\mathcal{F}(\mathbf{X})$, and each element in the filter is constrained in $(0, 1)$ after softmax. The proposed \method~then tunes a subspace coefficient $\boldsymbol{\alpha}\in \mathbb{R}^{H\times H}$ with a tiny number of parameters, that expands the range of attention scores by linearly combining the original filter subspace and enhances the expressiveness of the multi-head attention.}
    \label{fig:main_fig}
\vspace{-10pt}
\end{figure*}

We show that our approach can be seamlessly combined with previous PEFT methods in a plug-and-play manner. As the filter subspace in each attention layer is spanned by a tiny number of attention maps~(usually 10$\sim $30), we only introduce neglectable additional tunable parameters and training costs over the base PEFT methods.

We validate our approach on a large variety of fine-tuning tasks and datasets, along with pre-trained transformers of different scales and modalities. Specifically, we first show the effectiveness of our method with ViT~\cite{vit} on the few-shot image classification fine-tuning. We also show our method can achieve superior performance with vision-language transformers on image-text multi-modal fine-tuning~\cite{vlbart}. Moreover, we also show our method can better customize the text-to-image diffusion models on personalized generation~\cite{ruiz2023dreambooth}.

We summarize our contributions as follows,
\begin{itemize}
    
    \item We provide a filter subspace view of the multi-head attention and propose to fine-tune a new filter subspace by learning a small set of subspace coefficients.

    \item We demonstrate that our method can expand the feature space of the original multi-head attention and enhance its expressiveness on downstream tasks.
    
    \item We validate our method on extensive fine-tuning tasks and various transformer architectures.
\end{itemize}   
\section{Related Work}

\paragraph{Parameter-Efficient Fine-Tuning.}
Large attention-based transformers~\citep{vaswani2017attention, radford2021learning, rombach2022high, vit} gain a significant amount of generalizable knowledge by pre-training on large-scale datasets~\citep{ILSVRC15, schuhmann2022laion}. A common way to utilize those general skills is to fine-tune large transformers on domain-specific tasks and data.
Recent advancements in parameter-efficient fine-tuning (PEFT) methods aim to reduce the computational and memory costs of adapting large models by fine-tuning only a subset of parameters.
LoRA~\citep{hu2021lora} and its variants~\cite{kopiczko2023vera, karimi2021compacter, liu2022few, yeh2023navigating, zhang2023adalora, large_conv_ft} fine-tune lower-rank matrices at each layer to represent weight updates. DoRA~\cite{liu2024dora} enhances LoRA through weight decomposition into magnitude and direction, thus improving training stability and learning capacity without additional inference costs.
The adapter~\citep{adapter} approach inserts small modules between layers and reduces parameters by only tuning these adapters~\citep{chen2022adaptformer, karimi2021compacter, li2021prefix, pfeiffer2020adapterfusion, zaken2021bitfit}. 
Visual prompt tuning (VPT)~\citep{vpt, sohn2023visual} has introduced a limited number of learnable parameters for optimization while keeping the backbone frozen. 
SSF~\citep{ssf} proposes scaling and shifting features of each layer in a pre-trained large model.
OFT~\cite{oft} tackles fine-tuning in text-to-image diffusion models by preserving the hyperspherical energy of neuron relationships.
However, these methods primarily focus on efficient tuning of the parameter weight matrices, lacking a holistic view of the attention layer. In this paper, we take an orthogonal approach by tuning a tiny combination coefficient given a novel graph filter subspace view of the multi-head attention.


\vspace{-5pt}
\paragraph{The Expressiveness of Attention.}
Transformer-based models have achieved success across vision and language tasks~\cite{vlbart, vit, vqa, hudson2019gqa, lewis2019bart, liu2024visual} due to the effectiveness of the attention mechanism~\cite{vaswani2017attention}. Despite their success, various studies have examined the limitations in the expressiveness of transformers~\cite{dong2021attention, shi2022revisiting, wang2022anti, zhai2023stabilizing}. 
\citet{shi2022revisiting} and \citet{zhou2021deepvit} identify the attention collapse in transformers as a result of self-attention's convergence to a low-rank space, which leads to unstable training as attention scores concentrate~\cite{dong2021attention, noci2022signal}. To counteract this, techniques such as hierarchical fusion~\cite{shi2022revisiting} and stabilizing attention entropy~\cite{zhai2023stabilizing} have been proposed to prevent performance stagnation in vision transformers. DeepViT~\cite{zhou2021deepvit} proposes to re-generate diverse attention maps to avoid attention collapse in training vision transformers from scratch. In contrast, our work studies the expressiveness of the attention layer in the context of the large model fine-tuning. We also provide a novel interpretation of multi-head attention maps as a graph filter subspace, and propose the residual parameterization and regularization to stabilize the tuning process.

  
\section{Method}

\subsection{A Filter Subspace View on Attention}

\paragraph{Attention.}
The attention layer transforms the input sequence $\inp \in\mathbb{R}^{N\times C_i}$ to the output sequence $\out \in\mathbb{R}^{N\times C_o}$, where $N$ denotes the sequence length, $C_i$ and $C_o$ are the dimension of input and output features. 
The attention layer projects the input as $\Q = \inp \Wq$, $\K = \inp \Wk$, $\V = \inp \Wv$ using the corresponding projection matrices $\Wq, \Wk, \Wv \in \mathbb{R}^{C_i\times C_o}$, and calculates the attention map,
\begin{equation}
\begin{split}
    \attn &= \text{softmax}(\Q \K^{\intercal})\\
          &=\text{softmax}(\inp \Wq \Wk^{\intercal} \inp^{\intercal}).
\end{split}
\label{eq:attn_map}
\end{equation}
The attention map $\attn \in \mathbb{R}^{N \times N}$ captures the token-wise relationship by doing inner-product in a space transformed by $\Wq, \Wk$.
Then the output of the attention layer is,
\begin{equation}
\begin{split}
    \out & =\text{attn}(\Wq, \Wk, \Wv; \inp) \\
         & = \text{softmax}(\inp \Wq \Wk^{\intercal} \inp^{\intercal}) \inp \Wv \\
         & = \attn \inp \Wv.
\end{split}
\label{eq:attn}
\end{equation}

\paragraph{Multi-head Attention.}
Multi-head attention extends this by allowing multiple attention mechanisms, \textit{i.e.}, heads, to work in parallel, with each head independently learning attention patterns. For each attention head, we have,
\begin{equation}
    \out^{h} =\text{attn}(\Wq^{h}, \Wk^{h}, \Wv^{h}; \inp),
\label{eq:multi_head_attn}
\end{equation}
where $h=1,\cdots, H$ and $H$ is the number of heads. The project matrices for each head are $\Wq^h, \Wk^h, \Wv^h \in \mathbb{R}^{C_i\times \frac{C_o}{H}}$. 
The multi-head attention represents as,
\begin{equation}
\begin{split}
    \feat & = \text{Concat} \left[\out^{1}, \cdots, \out^{H}\right] \Wo \\
          & = \text{Concat} \left[\attn^1 \inp \Wv^1, \cdots, \attn^H \inp \Wv^H\right] \Wo \\
          & = \sum_{h=1}^{H} \attn^h \inp \Wvo^h,
\label{eq:multi_head_attn2}
\end{split}
\end{equation}
where $\Wo \in \mathbb{R}^{C_o\times C_o}$ and $\Wvo^h=\Wv^h (\Wo^h)^\intercal$, $\Wo^h \in \mathbb{R}^{C_o\times \frac{C_o}{H} }$, $\Wvo^h \in \mathbb{R}^{C_i \times C_o}$.


\vspace{-10pt}
\paragraph{A Filter Subspace View on Multi-head Attention.}
We now present our filter subspace view on the multi-head attention. By viewing each token embedding $\inp[n] \in \mathbb{R}^{C_i}$ as a node $V_n$, which are connected with all the other nodes with edges $E[m,n]=1$, $\forall m \in [N]$, we obtain a graph representation of all the tokens, $\mathcal{G}=(V, E)$. 

We then reformulate the attention operation as the message passing as in Graph Convolution Nerworks~\cite{kipf2016semi}. Specifically, inspired by \cite{l3net}, we write the single-head attention as a graph convolution with the filter $\filter^h(\inp)$,
\begin{equation}
\begin{split}
    \out^h[n, :] &= \sum_{\substack{n'\in [N] \\ c' \in [C_i]}} \filter^h(\inp)[n, n'] ~ \inp[n', c'] \W^h[c', :],
\end{split}
\label{eq:graph}
\end{equation}
where $\W^h=\Wvo^h \in \mathbb{R}^{C_i\times C_0}$ denotes the weight for linear projection on each node, and the $n$-th row of the filter, i.e., $\filter^h(\inp)[n, :]$, represents the local graph filter around the $n$-th node~(token). Comparing with (\ref{eq:attn_map}) and (\ref{eq:attn}), we have,
\begin{equation}
    \filter^h(\inp) = \text{softmax}(\inp \Wq \Wk^{\intercal} \inp^{\intercal}).
\end{equation}


The multi-head attention individually learns $H$ filters $\{\filter^1(X), ..., \filter^H(X)\}$ to aggregate complementary information across nodes. These $H$ filters naturally form a graph convolutional filter subspace $\mathcal{F}(\inp)$ as,
\begin{equation}
    \mathcal{F}(\inp) = \{\filter^h(\inp)\}_{h=1}^H \subseteq (0,1)^{N\times N},
\label{eq:filter_subspace}
\end{equation}
and we denote each $\filter^h(\inp)$ as a filter atom. 
The multi-head attention can then be represented in the way of decomposed graph convolution~\cite{l3net} as, 
\begin{equation}
\begin{split}
    \out[n,:] & = \sum_{h=1}^{H} \sum_{\substack{n'\in [N] \\ c' \in [C_i]}} \filter^h(\inp)[n, n'] \inp[n', c'] \W^h[c',:].
\end{split}
\label{eq:subspace_graph_conv}
\end{equation}

It can be seen from (\ref{eq:subspace_graph_conv}) that the parameters $\filter^h(\inp), \W^h$ plays two key yet independent roles in the graph convolution. The $\W^h$ updates the knowledge for each node by projecting its representation into a new space, while the graph filter $\filter^h(\inp)$ guides the information flow across nodes. Intuitively, with a more expressive graph convolutional filter subspace $\mathcal{F}(\inp)$, each node can aggregate more informative features from other nodes.

\subsection{Subspace Coefficient Tuning}
Given the re-formulation of multi-head attention maps as a graph convolutional filter subspace, we present an efficient fine-tuning method specifically designed for transformers to enhance attention's expressiveness. Our motivation is that after the softmax function, each graph filter $\filter^h(\inp)$ has values in the range $(0, 1)$ with row-sum as one. The normalized attention maps restrict the expressiveness in the way that each output embedding lies in the convex hull of the value embeddings.

To expand the feature space of the output embeddings, we propose \method, to tune the filter subspace on the target dataset with a small set of combination coefficients, i.e., \textbf{subspace coefficient} $\coeff \in \mathbb{R}^{H \times H}$. As shown in Figure~\ref{fig:main_fig}, the tunable coefficient $\coeff$ constructs a more expressive filter subspace $\hat{\mathcal{F}}_\alpha(\inp)$ by linearly combining the pre-trained multi-head attention maps $\mathcal{\filter}(\inp)$. Each element $\hat{\filter}^h(\inp)$ in the learned filter subspace can be written as,
\begin{equation}
    \hat{{\filter}}^h(\inp) = \coeff[h,:]^\intercal \mathcal{F}(\inp)=\sum_{i=1}^H \coeff[h, i] \filter^i(\inp).
\end{equation}
The multi-head attention then represents as,
\begin{equation}
\begin{split}
    \out & = \sum_{h=1}^{H} \hat{\filter}^h(\inp) \inp \W^h = \sum_{h=1}^{H} \sum_{i=1}^H \coeff[h, i] \filter^i(\inp) \inp \W^h.
\end{split}
\end{equation}
Note that, mathematically, the subspace coefficient $\coeff$ is unconstrained, allowing it to combine attention maps with a wider range of values from the original filter subspace $\mathcal{\filter}(\inp)$. With attention scores exceeding the range $(0, 1)$, we effectively expand the space for $\out$ by allowing a more flexible combination of value matrices. 

\vspace{-5pt}
\paragraph{Residual Parameterization.}To facilitate the tuning and preserve the most knowledge in the pre-trained transformer, we propose a residual parameterization of the subspace coefficient,
\begin{equation}
\begin{split}
    \coeff' &= \coeff + \mathbf{I}, \\
    \hat{{\filter}}^h(\inp) &= \filter^h(\inp) + \sum_{i=1}^H \coeff[h, i] \filter^i(\inp).
\end{split}
\label{eq:residual_param}
\end{equation}
In this way, we maintain the original filter subspace with the identity matrix, while parameterizing the subspace coefficient $\coeff'$ to learn the residual combination of attention maps on fine-tuning tasks. 

\vspace{-5pt}
\paragraph{Reguralization on Subspace Coefficient.}
The proposed \method~substantially enhances the expressiveness of multi-head attention with tunable coefficient $\coeff$. However, it may also lead to overfitting and preventing channel projection parameters $\Wv,\Wo$ from learning generalizable features. To alleviate this, we propose a simple regularization design that performs dropout~\cite{srivastava2014dropout} directly on the tuned subspace coefficient $\coeff$ during training,
\begin{equation}
    \coeff' = \text{Dropout}(\coeff; p) + \mathbf{I}.
\label{eq:dropout}
\end{equation}
In this way, we randomly remove an element from $\coeff$ with a probability of $p$ to avoid overfitting.

\begin{figure*}
    \centering
    \begin{subfigure}{0.98\textwidth}
        \includegraphics[trim={0pt 330pt 0pt 0pt}, clip, width=\textwidth]{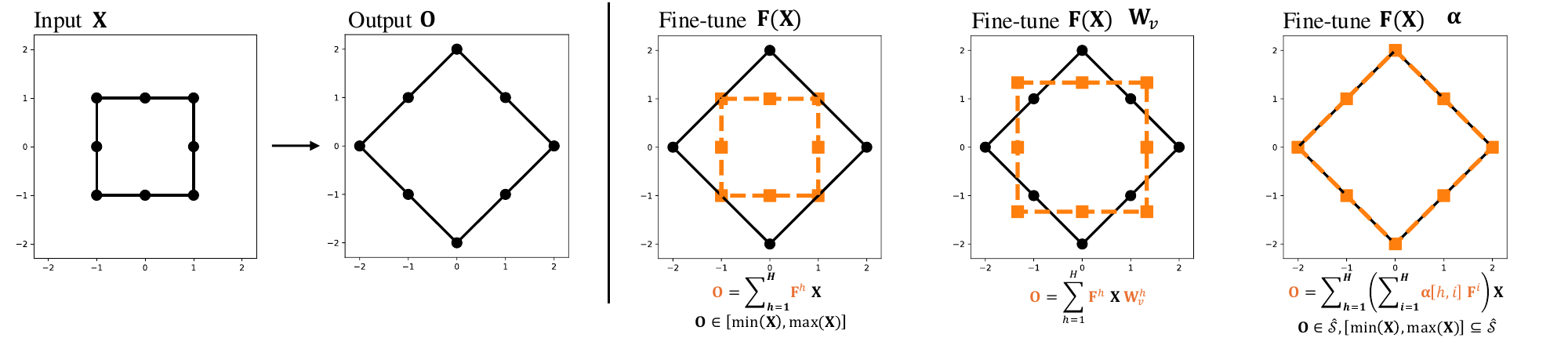}
    \end{subfigure}
    \caption{The effectiveness of tuning subspace coefficient $\coeff$. The goal is to convert the input graph $\inp$ to the target graph $\out$ using a single attention layer, $\out = \sum_{h=1}^{H} \filter^h(\inp) \inp \Wv^h$, where $\filter(\inp)=\text{softmax}(\inp \Wq \Wk^{\intercal} \inp^{\intercal})$. 
    We have the following observations: (i) When only tuning $\Wq$ and $\Wk$, \textit{i.e.}, updating $\filter(\inp)$, 
    the positions of output nodes are constrained within the range of the input nodes $\inp$, since the values of $\filter(\inp)$ are in the range of $(0,1)$. (ii) In addition to tuning $\filter(\inp)$, updating $\Wv$ projects the input as $\inp \Wv$, but the output graph is still not aligned with the target graph. (iii) 
    In addition to tuning $\filter(\inp)$, adapting $\coeff$ successfully transforms the input graph into the target graph. Compared with (i), $\coeff$ expand the positions of output nodes $\sum_{i=1}^{H}\coeff[:,i] \filter^i(\inp) \inp$ beyond the constraints of the input nodes, enhancing the expressiveness of the attention effectively.}
    \label{fig:toy}
\vspace{-10pt}
\end{figure*}

\section{Analysis}
In this section, we analyze that adjusting the filter subspace using the proposed subspace coefficient $\coeff$ can effectively enhance the expressiveness of the multi-head attention.

For the filter $\filter(\inp)$ in the filter subspace $\mathcal{F}$, since $\filter(\inp)=\text{softmax}(\inp \Wq \Wk^{\intercal} \inp^{\intercal})$, their elements $\filter(\inp)[n, n']$ satisfy the following properties, 
\begin{equation}
    0<\filter(\inp)[n, n']<1,~\sum_{n'\in [N]} \filter(\inp)[n, n'] = 1.
\end{equation}


\begin{definition}
    \label{prop:convex_set}
    Given $N$ nodes $\{\inpv_i\}_{i=1}^N$, the set $\mathcal{S}(\inpv) = \{ \sum_{i=1}^N \lambda_i \inpv_i \mid \lambda_i > 0, ~\sum_{i} \lambda_i = 1\}$ is a bounded convex set. 
\end{definition}
According to the Definition~\ref{prop:convex_set}, when the projected nodes $\inp \W^h \in \mathbb{R}^{N \times C_o}$ of each head $h$ are given, the output $\out^h[n,:] = \sum_{n'\in [N]} \filter^h(\inp)[n, n'] \inp[n'] \W^h$ lies within $\mathcal{S}(\inp \W^h)$, which means the outputs $\out^h[n,:]$ are bounded. This property also applies to the multi-head attention mechanism.

\begin{proposition}
\label{prop:mh_convex_set}
    If $\mathcal{S}^{h}$ is a bounded convex set, $\forall h=1,\cdots,H$, the set $\mathcal{S} = \mathcal{S}^{1}+\cdots+\mathcal{S}^{H}=\{\inpv^{1} + \cdots + \inpv^{H} \mid \forall \inpv^{h} \in \mathcal{S}^{h}\}$ is also a bounded convex set.
\end{proposition}
According to Proposition~\ref{prop:mh_convex_set}, when the projected nodes $\inp \W^h$ are given, the outputs of the multi-head attention $\out[n,:] = \sum_{h=1}^{H} \sum_{n'\in [N]} \filter^h(\inp)[n, n'] \inp[n'] \W^h$ are also bounded.

\begin{proposition}
    \label{prop:new_subspace}
    After applying the subspace coefficients $\coeff$, the outputs of the reconstructed filter subspace $\hat{\mathcal{F}}_\alpha$ form a set $\hat{\mathcal{S}} = \{ \sum_{h=1}^{H} \sum_{i=1}^{H} \coeff[h, i] \filter^i(\inp)[n, n'] \inp \W^h \mid \filter^i(\inp)[n, n'] > 0, ~\sum_{j} \filter^i(\inp)[n, n'] = 1, ~\coeff[h, i] \in \mathbb{R}\}$, which satisfies $\mathcal{S} \subseteq \hat{\mathcal{S}}$.
\end{proposition}
According to Proposition~\ref{prop:new_subspace}, when applying filter subspace coefficients $\coeff$, we can reconstruct the filter subspace $\mathcal{F}$ and expand the set $\mathcal{S}$. Specifically, given $\inp \W^h$, the output $\out[n,:] = \sum_{h=1}^{H} \sum_{i=1}^{H} \coeff[h, i] \filter^i(\inp)[n, n'] \inp \W^h$ can be in the set $\hat{\mathcal{S}} \setminus \mathcal{S}$. Our method obtains an expanded set, enhancing the expressiveness of the model. We provide the support of our analyses in Appendix~\ref{app:analysis}.


\subsection{Validation Experiment}
We assess the effectiveness of the subspace coefficients using a single attention layer. Given an input graph $\inp$, the output of the attention layer is calculated as $\out=\filter(\inp) \inp \Wv$, where $\filter(\inp)=\text{softmax}(\inp \Wq \Wk^{\intercal} \inp^{\intercal})$.

Our goal is to optimize $\Wq$, $\Wk$, $\Wv$ to transform the input graph $\inp$ into the target graph $\out$ as shown in Figure~\ref{fig:toy}. The input graph contains 8 nodes
$\inp = [[-1, 1],[0, 1],[1, 1],[1, 0],[1, -1],[0, -1],[-1, -1],[-1, 0] ]$ positioned in a range of $[-1, 1]$ and the target graph 
$\out = \begin{bmatrix} [-1, 1],[0, 2],[1, 1],[2, 0],[1, -1],[0, -2],[-1, -1],[-2, 0] \end{bmatrix}$, with node coordinates in the range of $[-2, 2]$. 

As shown in Figure~\ref{fig:toy}, by only tuning the $\Wq$ and $\Wk$, \textit{i.e.}, $\filter(\inp)=\text{softmax}(\inp \Wq \Wk^{\intercal} \inp^{\intercal})$, the output is constrained inside the range of the input, which is aligned with Proposition~\ref{prop:convex_set}.
When we tune $\Wq$, $\Wk$, and $\Wv$, the output exceeds the input range but still cannot fully match the target graph. However, when we tune the subspace $\filter(\inp)$ and subspace coefficient $\coeff$, the attention layer can successfully transform the input graph into the target graph as in Figure~\ref{fig:toy}. This observation is aligned with Proposition~\ref{prop:new_subspace}, and indicates that the proposed \method~can effectively expand the feature space of the multi-head attention layer.

\begin{table*}[t]
\caption{Fine-tuning Results on VTAB-1k with ViT} 
\centering
\vspace{-5pt}
\footnotesize
\scalebox{0.72}{
\begin{tabular}{c|ccccccc|cccc|cccccccc|ccc}
\toprule
                 & \multicolumn{7}{c|}{\textbf{Natural}} & \multicolumn{4}{c|}{\textbf{Specialized}} & \multicolumn{8}{c|}{\textbf{Structured}} &  \\ 
\midrule
                  & \rotatebox{90}{C100} & \rotatebox{90}{Cal101} & \rotatebox{90}{DTD} & \rotatebox{90}{F102} & \rotatebox{90}{Pets} & \rotatebox{90}{SVHN} & \rotatebox{90}{S397} & \rotatebox{90}{Came} & \rotatebox{90}{E\_SAT} & \rotatebox{90}{Res45} & \rotatebox{90}{Retino} 
                 & \rotatebox{90}{Cle\_cnt} & \rotatebox{90}{Cle\_dis} & \rotatebox{90}{DMLab} & \rotatebox{90}{KITTI}  & \rotatebox{90}{dS\_loc} & \rotatebox{90}{dS\_ori} & \rotatebox{90}{S\_azi} & \rotatebox{90}{S\_ele} & \rotatebox{90}{\normalsize \textbf{Mean}} & \rotatebox{90}{{\normalsize \textbf{Param.}}} & \rotatebox{90}{\footnotesize \textbf{GFLOPs}}\\
\midrule
Full fine-tuning & 68.9 & 87.7 & 64.3 & 97.2 & 86.9 & 87.4 & 38.8 
                 & 79.7 & 95.7 & 84.2 & 73.9
                 & 56.3 & 58.6 & 41.7 & 65.5 & 57.5 & 46.7 & 25.7 & 29.1
                 & \normalsize 65.57 & \normalsize 85.8 & 45.15 \\

\midrule
Adapter         & 74.1 & 86.1 & 63.2 & 97.7 & 87.0 & 34.6 & 50.8
                & 76.3 & 88.0 & 73.1 & 70.5
                & 45.7 & 37.4 & 31.2 & 53.2 & 30.3 & 25.4 & 13.8 & 22.1
                & \normalsize 55.82 & \normalsize 0.234 & 16.44\\
BitFit           & 72.8 & 87.0 & 59.2 & 97.5 & 85.3 & 59.9 & 51.4
                 & 78.7 & 91.6 & 72.9 & 69.8
                 & 61.5 & 55.6 & 32.4 & 55.9 & 66.6 & 40.0 & 15.7 & 25.1
                 &  \normalsize 62.05 & \normalsize 0.105 & 15.08\\
VPT              & 77.7 & 86.9 & 62.6 & 97.5 & 87.3 & 74.5 & 51.2 
                 & 78.2 & 92.0 & 75.6 & 72.9 
                 & 50.5 & 58.6 & 40.5 & 67.1 & 68.7 & 36.1 & 20.2 & 34.1
                 & \normalsize 64.85 & \normalsize 0.06 & 16.68\\
LoRA             & 67.1 & 91.4 & 69.4 & 98.8 & 90.4 & 85.3 & 54.0 
                 & 84.9 & 95.3 & 84.4 & 73.6 
                 & 82.9 & 69.2 & 49.8 & 81.7 & 81.8 & 48.3 & 32.8 & 44.2
                 & \normalsize 72.91 & \normalsize 0.305 & 15.74 \\
SPT-Adapter & 72.9 & 93.2 & 72.5 & 99.3 & 91.4 & 84.6 & 55.2
                & 85.3 & 96.0 & 83.9 & 75.8
                & 82.2 & 68.0 & 49.3 & 80.0 & 82.4 & 51.9 & 31.7 & 41.2
                & \normalsize 73.52 & \normalsize 0.266 & 16.44\\

\midrule
Linear Probing   & 63.4 & 85.0 & 63.2 & 97.0 & 86.3 & 36.6 & 51.0
                 & 78.5 & 87.5 & 68.6 & 74.0
                 & 34.3 & 30.6 & 33.2 & 55.4 & 12.5 & 20.0 & 9.6 & 19.2
                 & \normalsize 52.94 & \normalsize 0 & 15.05 \\

\cellcolor{lightgray!30}\textbf{Coeff. $\coeff$ Only (Ours)} 
& \cellcolor{lightgray!30}69.5 & \cellcolor{lightgray!30}90.8 & \cellcolor{lightgray!30}{74.5} & \cellcolor{lightgray!30}99.1 & \cellcolor{lightgray!30}90.9 & \cellcolor{lightgray!30}83.0 & \cellcolor{lightgray!30}51.1 & \cellcolor{lightgray!30}83.6 & \cellcolor{lightgray!30}95.3
& \cellcolor{lightgray!30}84.6 & \cellcolor{lightgray!30}{75.8}
& \cellcolor{lightgray!30}57.4 & \cellcolor{lightgray!30}59.1 & \cellcolor{lightgray!30}46.5 & \cellcolor{lightgray!30}78.6 & \cellcolor{lightgray!30}75.0 & \cellcolor{lightgray!30}51.4 & \cellcolor{lightgray!30}25.6 & \cellcolor{lightgray!30}34.2
& \cellcolor{lightgray!30}\normalsize \textbf{69.78} & \cellcolor{lightgray!30}\normalsize \textcolor{red}{0.002} & \cellcolor{lightgray!30}15.06\\



\midrule

FACT-TT$\leq$16   & 71.3 & 89.6 & 70.7 & 98.9 & 91.0 & 87.8 & 54.6 
                    & 85.2 & 95.5 & 83.4 & 75.7
                    & 82.0 & 69.0 & 49.8 & 80.0 & 79.2 & 48.4 & 34.2 & 41.4
                    & \normalsize 73.04 & \normalsize 0.040 & 15.14\\
\cellcolor{lightgray!30}\textbf{w/ Coeff-Tuning (Ours)} 
& \cellcolor{lightgray!30}\textbf{72.1} & \cellcolor{lightgray!30}\textbf{89.7} & \cellcolor{lightgray!30}\textbf{71.8} & \cellcolor{lightgray!30}\textbf{99.1} & \cellcolor{lightgray!30}\textbf{91.2} & \cellcolor{lightgray!30}\textbf{88.9} & \cellcolor{lightgray!30}\textbf{54.8} & \cellcolor{lightgray!30}\textbf{85.3} & \cellcolor{lightgray!30}\textbf{95.7}
& \cellcolor{lightgray!30}\textbf{84.2} & \cellcolor{lightgray!30}75.3
& \cellcolor{lightgray!30}\textbf{83.0} & \cellcolor{lightgray!30}68.8 & \cellcolor{lightgray!30}\textbf{51.0} & \cellcolor{lightgray!30}\textbf{81.3}
& \cellcolor{lightgray!30}\textbf{80.0} & \cellcolor{lightgray!30}\textbf{49.3} & \cellcolor{lightgray!30}\textbf{35.0} & \cellcolor{lightgray!30}\textbf{42.7} &  \cellcolor{lightgray!30}\normalsize \textbf{73.64} & \cellcolor{lightgray!30}\normalsize 0.042 & \cellcolor{lightgray!30}15.15\\

\midrule
SNELL-8 & 73.7 & 92.7 & 72.4 & 99.2 & 91.4 & 89.2 & 55.4
                & 84.9 & 96.1 & 86.4 & 75.2 
                & 84.0 & 68.5 & 53.5 & 81.0 & 82.7 & 49.9 & 33.9 & 39.2 
                & \normalsize 74.17 & \normalsize 0.268 & 15.47\\
\cellcolor{lightgray!30}\textbf{w/ Coeff-Tuning (Ours)}  
& \cellcolor{lightgray!30}\textbf{74.2} & \cellcolor{lightgray!30}\textbf{92.9} & \cellcolor{lightgray!30}\textbf{73.1} & \cellcolor{lightgray!30}99.2 & \cellcolor{lightgray!30}\textbf{91.6} & \cellcolor{lightgray!30}\textbf{90.3} & \cellcolor{lightgray!30}\textbf{55.6 }& \cellcolor{lightgray!30}\textbf{85.2} & \cellcolor{lightgray!30}\textbf{96.2}
& \cellcolor{lightgray!30}86.3 & \cellcolor{lightgray!30}\textbf{75.8}
& \cellcolor{lightgray!30}\textbf{84.1} & \cellcolor{lightgray!30}\textbf{68.8} & \cellcolor{lightgray!30}\textbf{53.7} & \cellcolor{lightgray!30}\textbf{81.3}
& \cellcolor{lightgray!30}\textbf{84.0} & \cellcolor{lightgray!30}\textbf{52.1} & \cellcolor{lightgray!30}\textbf{35.0} & \cellcolor{lightgray!30}\textbf{39.9} &  \cellcolor{lightgray!30}\normalsize \textbf{74.70} & \cellcolor{lightgray!30}\normalsize 0.270 & \cellcolor{lightgray!30}15.48\\

\bottomrule
\end{tabular}
}
\label{tab:discriminative_vtab_main}
\end{table*}

\subsection{Complexity Analysis}
We now formally analyze the complexity of the proposed \method, and compare it with other baseline PEFT methods. The number of parameters required for our method is $H \times H$, where $H$ is the number of heads at each attention layer. Considering $\Wq, \Wk, \Wv \in \mathbb{R}^{C_i\times C_o}$ and $\Wo \in \mathbb{R}^{C_o\times C_o}$ of the attention layer, our method introduces relative $\frac{H^2}{3 C_i C_o + C_o^2}$ parameters. In modern large transformer models, we typically have $C_i=2048$ and $C_o=1280$. With $H=10$, our method requires approximately $0.001\%$ additional parameters. Compared with other PEFT methods such as LoRA, our method introduces relative $\frac{H^2}{3 r C_i + 5 r C_o}$ parameters, where $r$ is the LoRA rank. Using the same values for $C_i$, $C_o$, $H$, and setting $r=4$, our approach requires less than $0.2\%$ additional parameters compared to LoRA and its variants. This also indicates that our \method~can maintain almost the same quantity of parameters when combined with other PEFT methods.


\section{Experiments}
In this section, we validate the proposed \method~with various tasks and datasets, including tuning the vision transformer~\cite{vit} on few-shot image classification, tuning text-to-image diffusion model~\cite{podell2023sdxl} for concept customization, and tuning vision-language BART~\cite{vlbart} on image-to-text understanding. Combining with other weight-based PEFT methods, the proposed \method~consistently achieves superior results with neglectable parameters added.

\begin{figure*}[t]
    \centering
    \includegraphics[trim={90pt 60pt 75pt 40pt}, clip, width=0.9\linewidth]{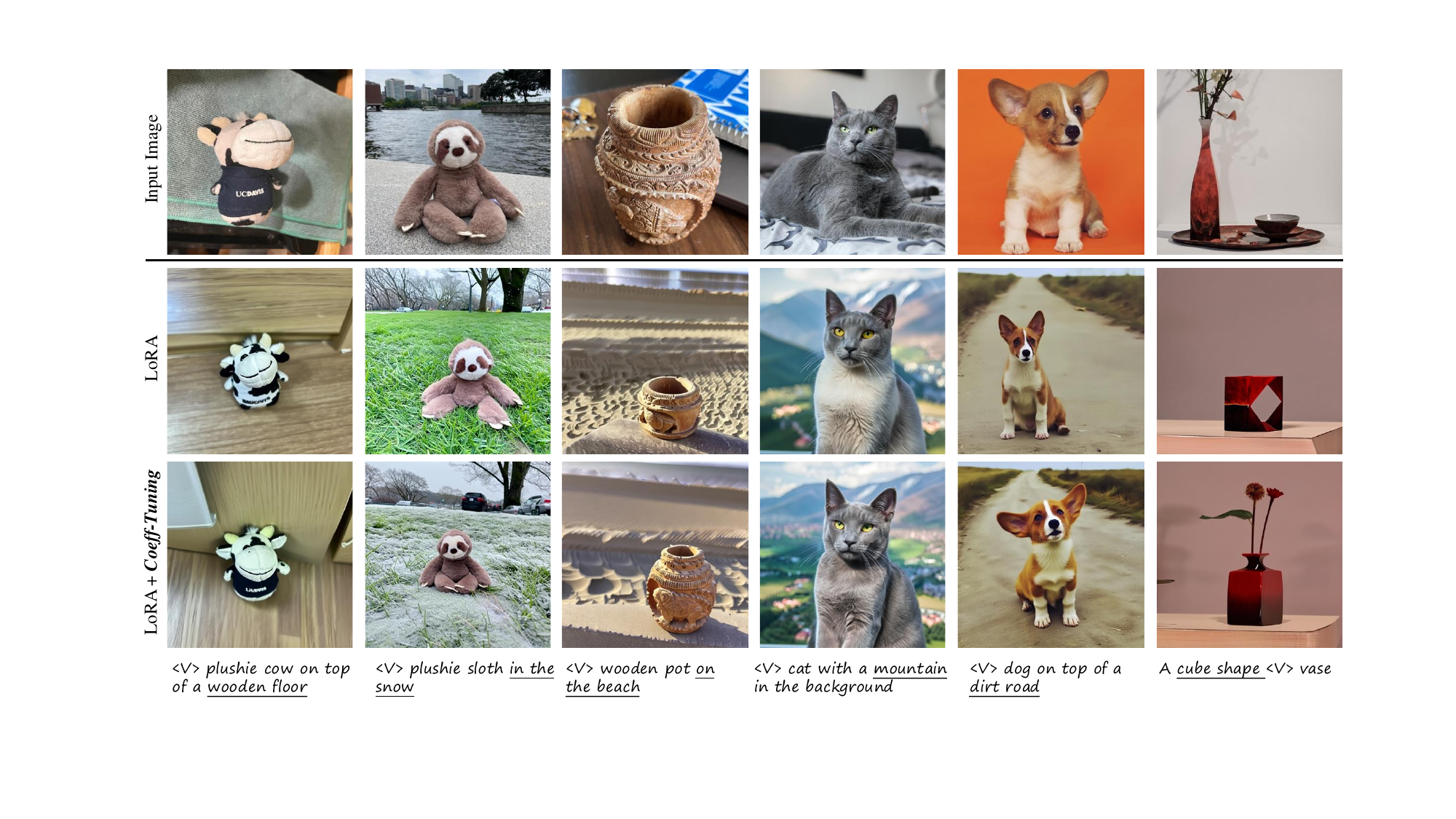}
    \vspace{-5pt}
    \caption{Results on Concept Customization. Compared to the baseline method (middle row), \method~(bottom row) leads to higher concept fidelity and text alignment of the generated images. For instance, our method accurately generates images of a ``$\langle V \rangle$ plushie sloth in the snow," while the baseline fails to align with the text prompt. Additionally, our method generates a “$\langle V \rangle$ wooden pot” with higher fidelity to the input image.}
    \label{fig:db}
    \vspace{-7pt}
\end{figure*}

\subsection{Few-shot Image Classification}
\label{sec:vit_tuning}
we first validate the proposed method by fine-tuning Vision Transformers on a series of downstream datasets for few-shot image classification. Our method achieves superior results while maintaining a high parameter efficiency compared with baseline methods.

\begin{table}[b]
\vspace{-15pt}
\caption{\footnotesize Fine-tuning Results on the FGVC with ViT.}
\vspace{-10pt}
\centering
\scalebox{0.64}{
\hspace{-15pt}
\begin{tabular}{c|ccccc|c}
\toprule
                     & CUB-200 & NABirds & Flowers & S-Dogs & S-Cars & Mean \\ \midrule
Full fine-tuning     & 87.3    & 82.7    & 98.8           & 89.4          & 84.5          & 88.5      \\
Linear Probing       & 85.3    & 75.9    & 97.9           & 86.2          & 51.3          & 79.3      \\ \midrule
Adapter              & 87.3    & 84.3    & 98.4           & 88.8          & 68.4          & 85.5      \\
BitFit               & 88.4    & 84.2    & 98.8           & 91.2          & 79.4          & 88.4      \\
VPT                  & 86.7    & 78.8    & 98.4           & 90.7          & 68.7          & 84.6      \\
LoRA                 & 84.9    & 79.0    & 98.1           & 88.1          & 79.8          & 86.0      \\
SPT-LoRA             & 88.6    & 83.4    & 99.5           & 91.4          & 87.3          & 90.1      \\ \midrule
SNELL-8              & 89.6    & 86.8    & 99.3           & 92.1          & 89.9          & 91.5      \\
\cellcolor{lightgray!30}\textbf{w/ Coeff-Tuning (Ours)}              & \cellcolor{lightgray!30}\textbf{89.8}    & \cellcolor{lightgray!30}\textbf{87.1}    & \cellcolor{lightgray!30}\textbf{99.5}           & \cellcolor{lightgray!30}\textbf{92.2}          & \cellcolor{lightgray!30}\textbf{90.7}          & \cellcolor{lightgray!30}\textbf{91.9}      \\
\bottomrule
\end{tabular}
}
\label{tab:discriminative_fgvc_main}
\end{table}
\vspace{-10pt}

\paragraph{Experimental Settings.}
We utilize the VTAB-1k~\cite{vtab} dataset which comprises 19 diverse computer vision tasks across three domains: standard natural imagery, specialized capture, and synthetic/structured environments. The benchmark evaluates model performance on various task objectives. With only 1,000 training samples per task, it presents a significant few-shot learning challenge. We also benchmark our method on five fine-grained datasets in few-shot settings, including CUB-200-2011~\cite{cub}, NABirds~\cite{nabird}, Oxford Flowers~\cite{flower}, Stanford Dogs~\cite{s_dog} and Stanford Cars~\cite{s_car}. A detailed description is provided in Appendix~\ref{app:exp_vit}.

As for the vision transformer, we follow VPT~\cite{vpt}, SSF~\cite{ssf}, and select ViT-B/16~\cite{vit} pre-trained on ImageNet-21K~\cite{deng2009imagenet} as the base model for fine-tuning. We benchmark the proposed \method ~with a series of PEFT methods for few-shot learning including SSF~\cite{ssf}, VPT~\cite{vpt}, Adapter~\cite{adapter}, LoRA~\cite{hu2021lora}, and BitFit~\cite{zaken2021bitfit}. 

We present two settings of \method~in this task. We first explore the representation ability of \textbf{tuning subspace coefficient $\coeff$ only}, which tunes the transformer bottleneck with only tuning $H^2$ (\textbf{144} for ViT~\cite{vit}) parameters for each transformer block. We also show the performance of combining \method~and SSF~\cite{ssf} to demonstrate that our method can plug and play with other PEFT methods. We set the coefficient dropout rates as $p=0$ and $p=0.1$ for these two settings respectively, and provide other hyperparameters in Appendix~\ref{app:exp_vit}.

\vspace{-5pt}
\paragraph{Results.}
As shown in Table~\ref{tab:discriminative_vtab_main}, the proposed \method~shows superior performances in both settings. Compared with the linear probing baseline which achieves a mean accuracy of 61.69, \method~obtains a significantly higher result of \textbf{73.49} with tuning only $\mathbf{(12*12)*12=1,728}$ additional subspace coefficients $\coeff$ in ViT~\cite{vit} (ViT~\cite{vit} has 12 attention layers, each attention has 12 heads). Thus, we achieve a large performance increase in few-shot image classification~(\textbf{11.8}$\mathbf{\uparrow}$) with neglectable additional parameter costs. Note that this result has already surpassed almost all baselines. We further achieve the best performance by combining \method~with SSF while maintaining almost the same parameter cost.

\subsection{Personalized Text-to-Image Generation}

We now show the effectiveness of \method~with the concept customization with Stable Diffusion (SDXL)~\cite{podell2023sdxl, rombach2022high}. We validate the effectiveness of \method~by combining it with baseline PEFT methods, LoRA~\cite{hu2021lora} and DoRA~\cite{liu2024dora}, and compare the performance with baselines without tuning $\coeff$. Specifically, we fine-tune SDXL~\cite{podell2023sdxl} to associate the special token with a concept image from Dreambooth~\cite{ruiz2023dreambooth} and CustomConcept101~\citep{kumari2023multi} datasets.  We choose 10 different concepts, and fine-tune the model on $4\sim 9$ images. We utilize 25 different text prompts following~\cite{oft} and generate 4 images for each prompt. We evaluate the fidelity, diversity, and text-to-image alignment of the generated images. Other training details are described in Appendix~\ref{app:exp_dreambooth}.

\begin{table}[t]
    \centering
    \caption{Concept Customization Results with SDXL.}
    \vspace{-5pt}
    \resizebox{.48\textwidth}{!}{
    \begin{tabular}{l|c|c|c|c}
    \toprule
            Method & \# Params (\%) & Text Align. & Concept Align. & Diversity \\
    \midrule
        DoRA & 0.0371 & 0.33 & 0.73 & 10.9\\
        \cellcolor{lightgray!30}\textbf{Ours + DoRA} & \cellcolor{lightgray!30}0.0374 & \cellcolor{lightgray!30}\textbf{0.34} & \cellcolor{lightgray!30}\textbf{0.75} & \cellcolor{lightgray!30}\textbf{11.37}\\ 
    \midrule
    \midrule
        LoRA & 0.0306 & 0.33 & 0.74 & 10.08 \\ 
        \cellcolor{lightgray!30}\textbf{Ours + LoRA} & \cellcolor{lightgray!30}0.0309 & \cellcolor{lightgray!30}\textbf{0.34} & \cellcolor{lightgray!30}\textbf{0.75} & \cellcolor{lightgray!30}\textbf{10.45}\\
    \bottomrule
    \end{tabular}
    }
    \label{tab:db}
\vspace{-10pt}
\end{table}

    

\begin{table*}[t]
    \centering
    \caption{Multi-Model Fine-tuning Results with VL-Bart.}
    \vspace{-5pt}
    \resizebox{0.6\linewidth}{!}{
    \begin{tabular}{l|c|c|c|c|c|c}
    \toprule
        Method & \# Params ($\%$) & VQA$^{\text{v2}}$ & GQA & NVLR$^2$ & COCO Cap & Avg. \\
    \midrule
    Full Model & 100 & 66.9 & 56.7 & 73.7 & 112.0 & 77.3 \\
    \midrule
    \midrule
    LoRA & 5.94 & 65.2 & 53.6 & 71.9 & \textbf{115.3} & 76.5 \\ 
    \cellcolor{lightgray!30}\textbf{Ours + LoRA} & \cellcolor{lightgray!30}5.94 & \cellcolor{lightgray!30}\textbf{66.2} & \cellcolor{lightgray!30}\textbf{54.1} & \cellcolor{lightgray!30}\textbf{73.0} & \cellcolor{lightgray!30}115.2 & \cellcolor{lightgray!30}\textbf{77.1} \\
    \midrule
    \midrule
    DoRA & 5.96 & 65.8 & 54.7 & \textbf{73.1} & 115.9 & 77.4 \\
    \cellcolor{lightgray!30}\textbf{Ours + DoRA} & \cellcolor{lightgray!30}5.96 & \cellcolor{lightgray!30}\textbf{66.4} & \cellcolor{lightgray!30}\textbf{54.9} & \cellcolor{lightgray!30}73.0 & \cellcolor{lightgray!30}\textbf{116.4} & \cellcolor{lightgray!30}\textbf{77.7} \\
    \bottomrule
    \end{tabular}
    }
    \label{tab:vlbart}
    \vspace{-7pt}
\end{table*}

\vspace{-5pt}
\paragraph{Evaluation metrics.} 
We assess the concept alignment with the average cosine similarity between CLIP embeddings~\citep{radford2021learning} of the generated images and the concept images. The diversity of generated images is measured by the Vendi score~\citep{friedman2022vendi} calculated with the DINOv2 embeddings~\citep{oquab2023dinov2}. The text alignment of the generated images is measured via average cosine similarity in the CLIP feature space~\citep{radford2021learning}.

\vspace{-10pt}
\paragraph{Results.} 
The quantitative results are shown in Table~\ref{tab:db}.
By combining with LoRA~\cite{hu2021lora} or DoRA~\cite{liu2024dora}, the proposed \method~generates images with greater diversity and better alignment with the both text prompt and input concept images.
The qualitative results of our method compared with LoRA~\cite{hu2021lora} are shown in Figure~\ref{fig:db}. The generated images with \method~have higher concept fidelity, preserving more characteristics from the input images. 

As an example, consider the concept ``$\langle V \rangle$ cat" as shown in the 4th column in Figure~\ref{fig:db}, the image generated using our method retains the cat's color from the input image, while the one generated by the baseline method alter the color. Additionally, the backgrounds in images generated with our method show more detailed features compared to those from the baseline method. \method~achieves the above improvements by introducing a limited number of additional parameters. We provide more generated images in Appendix~\ref{app:exp_dreambooth}.

\begin{table}[b]
    \centering
    \caption{Ablation studies on (a) residual parameterization and (b) coefficient dropout showing their effects on model performance.}
    \begin{subtable}[b]{0.5\linewidth}
        \centering
        \small
        \begin{tabular}{c|c}
        \toprule
            Parametrization & Avg. \\
        \midrule
            $\coeff$ w/ rand. init. & 44.28 \\
            $\coeff$ w/ identity init. & 61.57 \\
            \cellcolor{lightgray!30}\textbf{$I + \coeff$, zero init.} & \cellcolor{lightgray!30}\textbf{73.49}\\
        \bottomrule
        \end{tabular}
        \caption{}
        \label{tab:ablation_residual}
    \end{subtable}
    \hspace{2pt}
    \begin{subtable}[b]{0.4\linewidth}
        \centering
        \small
        \begin{tabular}{c|c}
        \toprule
            Dropout $p$ & Avg. Score \\
        \midrule
            0.0 & 75.1 \\
            0.1 & 77.3 \\
            \cellcolor{lightgray!30}\textbf{0.2} & \cellcolor{lightgray!30}\textbf{77.7} \\
            0.4 & 72.3 \\
        \bottomrule
        \end{tabular}
        \caption{}
        \label{tab:ablation_dropout}
    \end{subtable}
    \label{fig:ablation_studies}
\end{table}

\subsection{Image-Text Understanding}
\label{sec:vlbart}
As we have shown the effectiveness of \method~in fine-tuning models on both few-shot image classification and personalized image generation, we now validate it on multi-task image-text understanding, which is a much larger-scale multi-modal fine-tuning task that requires a higher model capacity. By combining with baseline PEFT methods, our method again achieves superior performance with neglectable parameter increase.

\paragraph{Experimental Settings.}
We evaluate our method on VL-BART~\cite{vlbart} multi-modal transformer, which utilizes CLIP-ResNet101~\cite{radford2021learning} for vision encoding and sends the visual features to BART-Base~\cite{lewis2019bart} for image-text joint modeling. Following \cite{vladapter}, we evaluate our method with the multi-task fine-tuning of VL-Bart~\cite{vlbart} on four large-scale image-text understanding tasks simultaneously: VQAv2~\cite{vqa} and GQA~\cite{hudson2019gqa} for visual question answering, NLVR2~\cite{nlvr} for visual reasoning, and MSCOCO~\cite{chen2015microsoft} for image captioning. Following the plug-and-play practice, we combine the proposed \method~with LoRA~\cite{hu2021lora} and DoRA~\cite{liu2024dora} respectively, and set the subspace coefficient dropout rate to $p=0.2$. Detailed descriptions of other settings and hyperparameters are provided in Appendix~\ref{app:exp_image_text}.

\vspace{-5pt}
\paragraph{Results.}
As shown in Table~\ref{tab:vlbart}, our method surpasses baseline LoRA~\cite{hu2021lora} and DoRA~\cite{liu2024dora} respectively in almost all image-text datasets consistently, and achieves higher average scores. The \method~combined with LoRA~\cite{hu2021lora} obtains comparable results with full-model fine-tuning, and it achieves better results than full-model tuning when it is combined with DoRA~\cite{liu2024dora}. Notably, these improvements are achieved while maintaining the same parameter budgets as we only tune a hundred parameters for each attention.

\subsection{Ablation Study}
In this section, we examine the effectiveness of each component of our design with ablation experiments. 

\paragraph{Residual Parameterization.} 
We first demonstrate the importance of residual parameterization in Eq.~\ref{eq:residual_param} in \method. We compare our residual parameterization with the direct parameterization where we directly tune $\coeff$ without adding it to the identity matrix, and initialize it with either a random matrix or an identity matrix. We adopt the few-shot tuning task as in Sec.~\ref{sec:vit_tuning} and only tune $\coeff$. Results in Table~\ref{tab:ablation_residual} show that models with residual parameterization achieve superior performance than direct parameterization with two different initializations. This validates our design that maintaining the original filter subspace while learning the residual combination helps preserve pre-trained knowledge and leads to better fine-tuning performance.


\vspace{-5pt}
\paragraph{Subspace Coefficient Dropout.} 
The dropout regularization on the subspace coefficient as in Eq.~\ref{eq:dropout} plays a crucial role in preventing overfitting. We adopt the experiment setting in multi-modal fine-tuning as in Sec.~\ref{sec:vlbart}. We experiment with different dropout rates $p\in \{0, 0.1, 0.2, 0.4\}$ during training. The results in Table~\ref{tab:ablation_dropout} demonstrate that a moderate dropout rate ($p=0.2$) yields the best performance while either a lower or higher dropout rate leads to sub-optimal results. This suggests that applying dropout directly on the tuned subspace coefficient $\coeff$ can effectively prevent overfitting and lead to better generalization.

\section{Conclusion}
In this paper, we presented \method, a new parameter-efficient fine-tuning approach that enhances multi-head attention's expressiveness by learning a set of subspace coefficients. By reformulating attention as a graph convolution with a filter subspace, we demonstrated both theoretically and empirically that our method expands the feature space of multi-head attention beyond the original convex hull constraints. The proposed \method~can be seamlessly integrated with existing PEFT methods while introducing negligible additional parameters. Through extensive experiments across diverse tasks and modalities, we consistently demonstrated superior performance compared to competitive PEFT baselines. Our method achieved significant improvements in various metrics while maintaining high parameter efficiency, suggesting a promising direction for enhancing the adaptability of large pre-trained transformers.

\newpage
\clearpage
{
    \small
    \bibliographystyle{ieeenat_fullname}
    \bibliography{dcf_attention}
}

\renewcommand{\thesection}{\Alph{section}}
\setcounter{section}{0}

\clearpage
\setcounter{page}{1}
\maketitlesupplementary

\section{More on Analysis}
\label{app:analysis}
\begin{proposition}
    Given $N$ nodes $\{\inpv_i\}_{i=1}^N$, the set $\mathcal{S}(\inpv) = \{ \sum_{i=1}^N \lambda_i \inpv_i \mid \lambda_i > 0, ~\sum_{i} \lambda_i = 1\}$ is a bounded convex set. 
\end{proposition}

\begin{proof}
    For each element $\inpv'=\sum_{i=1}^N \lambda_i \inpv_i$ of $\mathcal{S}(\inpv)$, we have,
    \begin{equation}
        \left\lVert \inpv' \right\rVert = \left\lVert \sum_{i=1}^N \lambda_i \inpv_i \right\rVert 
                                        \leq \sum_{i=1}^N \lambda_i \left\lVert  \inpv_i \right\rVert.
    \end{equation}
    With given $\{\inpv_i\}_{i=1}^N$, every element $\inpv'=\sum_{i=1}^N \lambda_i \inpv_i$ is bounded. Therefore, $\mathcal{S}(\inpv)$ is a bounded convex set.
\end{proof}

\begin{proposition}
    If $\mathcal{S}^{h}$ is a bounded convex set, $\forall h=1,\cdots,H$, the set $\mathcal{S} = \mathcal{S}^{1}+\cdots+\mathcal{S}^{H}=\{\inpv^{1} + \cdots + \inpv^{H} \mid \forall \inpv^{h} \in \mathcal{S}^{h}\}$ is also a bounded convex set.
\end{proposition}

\begin{proof}
    For each element $\inpv'= \sum_{h=1}^{H} \inpv^h$ of $\mathcal{S}$, we have,
    \begin{equation}
        \left\lVert \inpv' \right\rVert = \left\lVert \inpv^1+\cdots+\inpv^H \right\rVert
                                        \leq \left\lVert \inpv^1\right\rVert + \cdots + \left\lVert \inpv^H\right\rVert.
    \end{equation}
    Since $\left\lVert \inpv^h\right\rVert, \forall h=1,\cdots,H$ are bounded, $\left\lVert \inpv' \right\rVert$ is also bounded. Therefore, $\mathcal{S}=\mathcal{S}^{1}+\cdots+\mathcal{S}^{H}$ is also a bounded convex set.
\end{proof}

\paragraph{Detailed calculation of parameters.}
We have provided the analysis and estimation of the parameters in Sec. 4.2 and Sec. 5 in the submission. Our method requires a significantly small number of parameters. In ViT for example, our method only requires an additional $\mathbf{12\times12\times12=0.0017}$\textbf{M} parameters~(12 heads with 12 layers), which is negligible. We provide the number of parameters with more precisions in Table~\ref{tab:discriminative_vtab_main}.

\paragraph{Complexity Analysis.}
We provide comparisons of training flops of forward pass \& backward pass of a single image in Table~\ref{tab:discriminative_vtab_main}. Our method utilizes comparable training flops with other baseline methods while achieving consistent performance improvement.

\section{Supplementary Experimental Details}
\subsection{Task and Experiment Backgrounds}
\paragraph{More on PEFT.}
Besides the widely-adapted LoRA~\cite{hu2021lora} and following works~\cite{liu2024dora}, \cite{large_conv_ft} propose the filter subspace decomposition for weight matrices. The filter subspace decomposition method~\cite{qiu2018dcfnet} has shown effectiveness in continual learning~\cite{miao2021continual,chen2024inner,wang2021learning}, video representation learning~\cite{miao2021spatiotemporal}, graph learning~\cite{cheng2021graph}, and generative tasks~\cite{wang2021image, wang2019stochastic, wang2021adaptive,li2024cumulative}.
There are some other works on fine-tuning the SVD decomposition of weights\cite{han2023svdiff}, Kronecker decomposition~\cite{patel2024efficient, patel2025learning, patel2023learning}, sparsity~\cite{wang2024roselora,liu2025unlocking, li2024training}, non-linearity~\cite{zhong2024neat} or fine-tunig bias parameters\cite{xie2023difffit}.

\paragraph{Tuning Vision Foundation Models.}
As there are emergent needs to customize the pre-trained foundation models for downstream tasks, a large corpus of fine-tuning methods has been proposed for both both discriminative and generative tasks. Among them, \cite{ruiz2023dreambooth, miao2024training, miao2024tuning, xiong2024groundingbooth,tarres2024thinking, song2024imprint, song2024refine, he2023diffusion, he2023globalmapper, he2024coho, he2024kubrick} have focused on tuning pre-trained image diffusion models for personalized generation, diversity, compositional generation, and human preference. While \cite{vpt,karimi2021compacter} propose to tune propose to tune vision transformers for downstream discriminative tasks.

\subsection{ViT Fine-tuing}
\label{app:exp_vit}
We first describe our selection of high-resolution sub-tasks from the 19 VTAB~\cite{vtab} fine-tuning dataset. Specifically, we select datasets containing images with a resolution equal to or higher than 224, corresponding to the pre-training image in ImageNet-21k~\cite{deng2009imagenet}. The selected datasets are shown in Table~\ref{tab:discriminative_vtab_main}.

We now present the training details. For the first setting, i.e., tuning subspace coefficient $\coeff$ only, we only add for each attention layer a subspace coefficient $\coeff$ with the proposed parameterization and tune $\coeff$ together with the linear head for each task. We set the dropout rate $p=0$. For the second case, we add $\coeff$ in the same way while adding the scaling and shift parameters as in SSF~\cite{ssf}, and set the dropout rate $p=0.1$. For both settings, we adopt the batch size of 64 and train the model on each task with the AdamW optimizer for 100 epochs.

\subsection{Concept Customization}
\label{app:exp_dreambooth}
In this experiment, we choose 10 concepts from Dreambooth~\cite{ruiz2023dreambooth} and Custom Diffusion~\cite{kumari2023multi}. These concepts include toys, objects, and animals. We generate images with 25 text prompts adapted from Dreambooth~\cite{ruiz2023dreambooth}. We utilize Adam~\cite{kingma2014adam} optimizer with a learning rate of $3 \times 10^{-4}$ and fine-tune the SDXL for 200 steps. The ranks of LoRA~\cite{hu2021lora} and Dora~\cite{liu2024dora} are $r=2$. We generate 4 different images with the shape of $1024 \times 1024$ for each text prompt. 

\begin{figure*}[t]
    \centering
    \includegraphics[trim={90pt 60pt 75pt 40pt}, clip, width=0.98\linewidth]{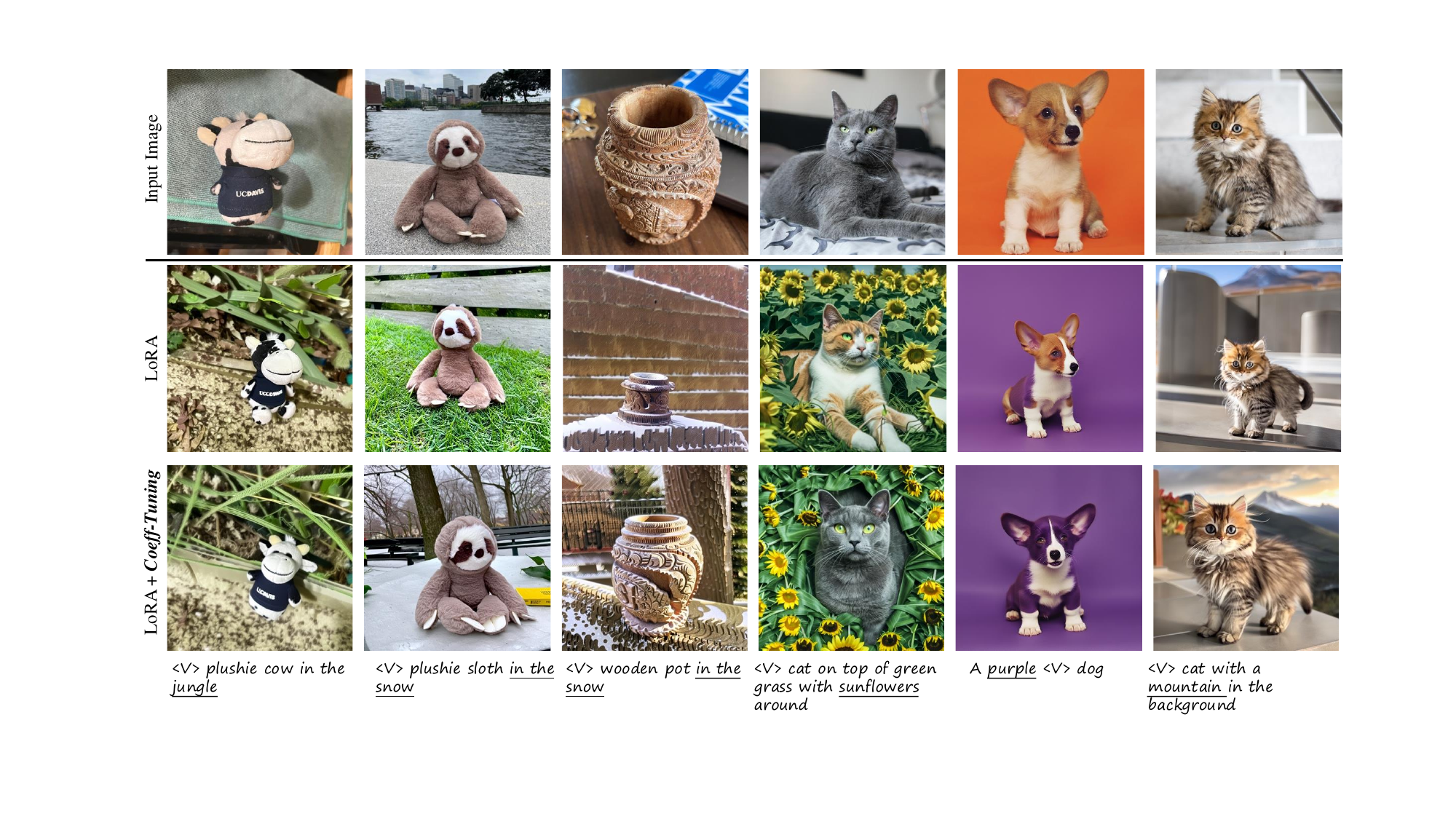}
    \caption{Results on Concept Customization.}
    \label{fig:db_1}
\end{figure*}

\begin{figure*}[t]
    \centering
    \includegraphics[trim={90pt 60pt 75pt 40pt}, clip, width=0.98\linewidth]{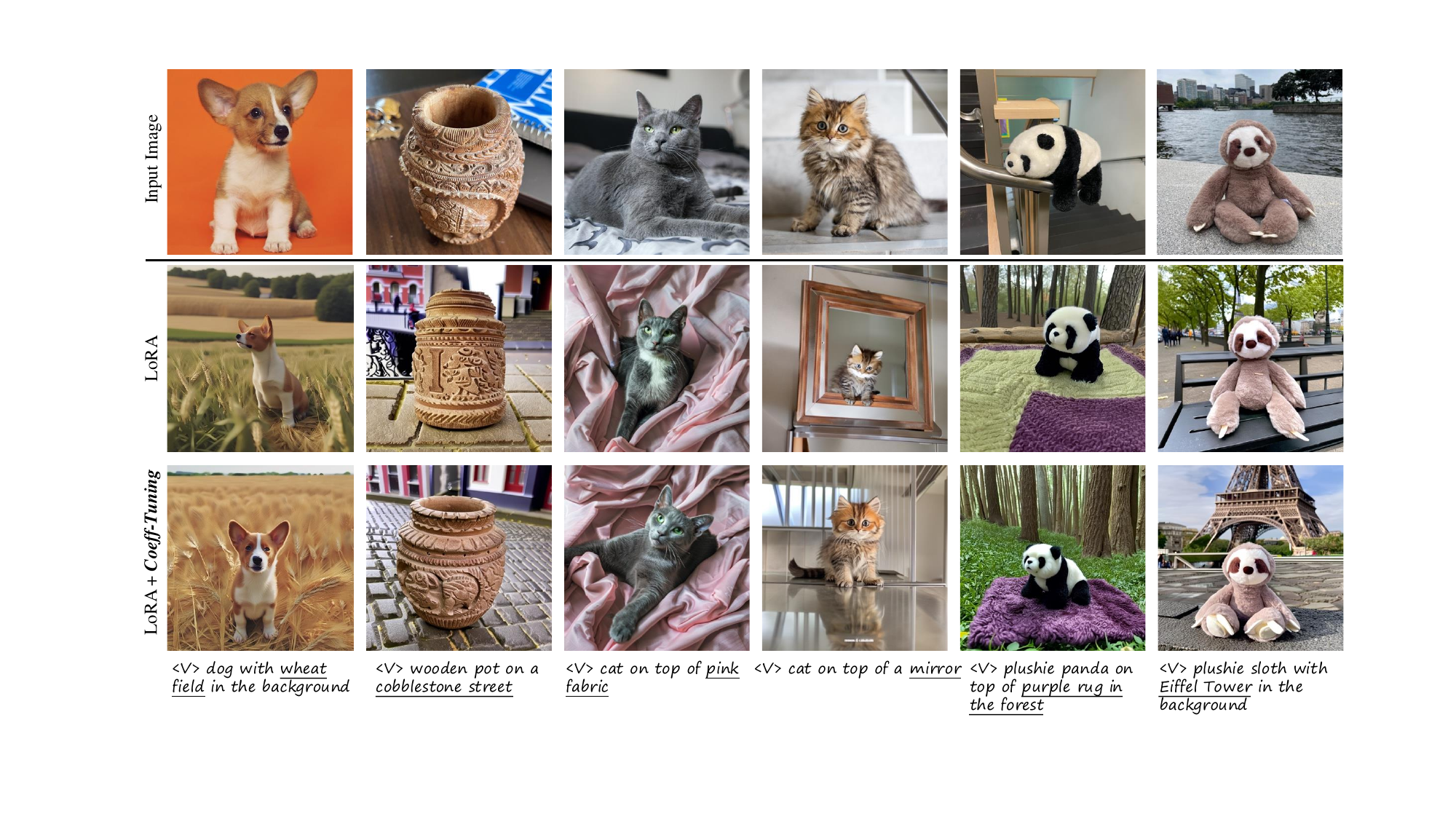}
    \caption{Results on Concept Customization.}
    \label{fig:db_2}
\end{figure*}

\begin{figure*}[t]
    \centering
    \includegraphics[trim={90pt 60pt 75pt 40pt}, clip, width=0.98\linewidth]{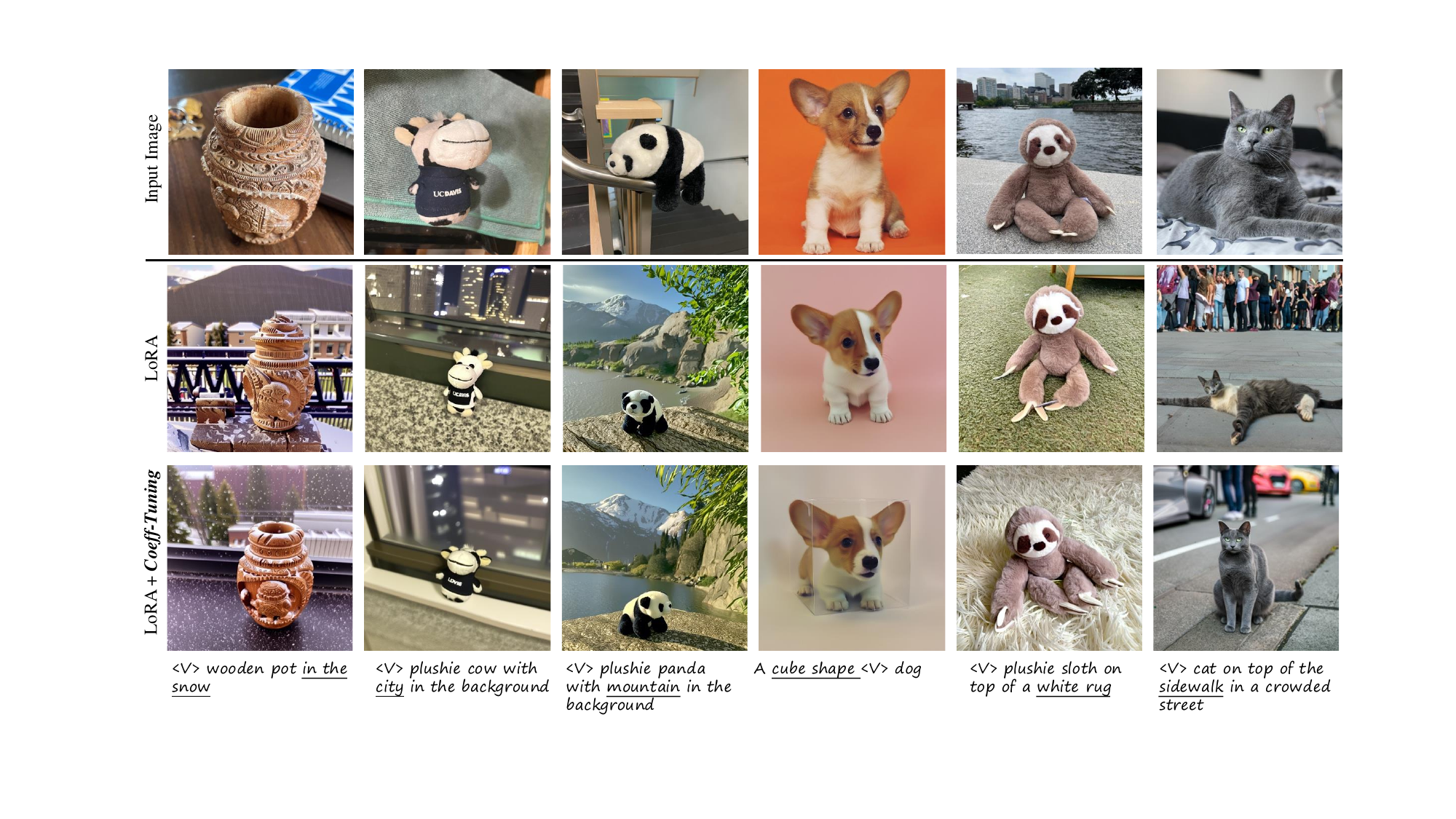}
    \caption{Results on Concept Customization.}
    \label{fig:db_3}
\end{figure*}

\begin{figure*}[t]
    \centering
    \includegraphics[trim={90pt 60pt 75pt 40pt}, clip, width=0.98\linewidth]{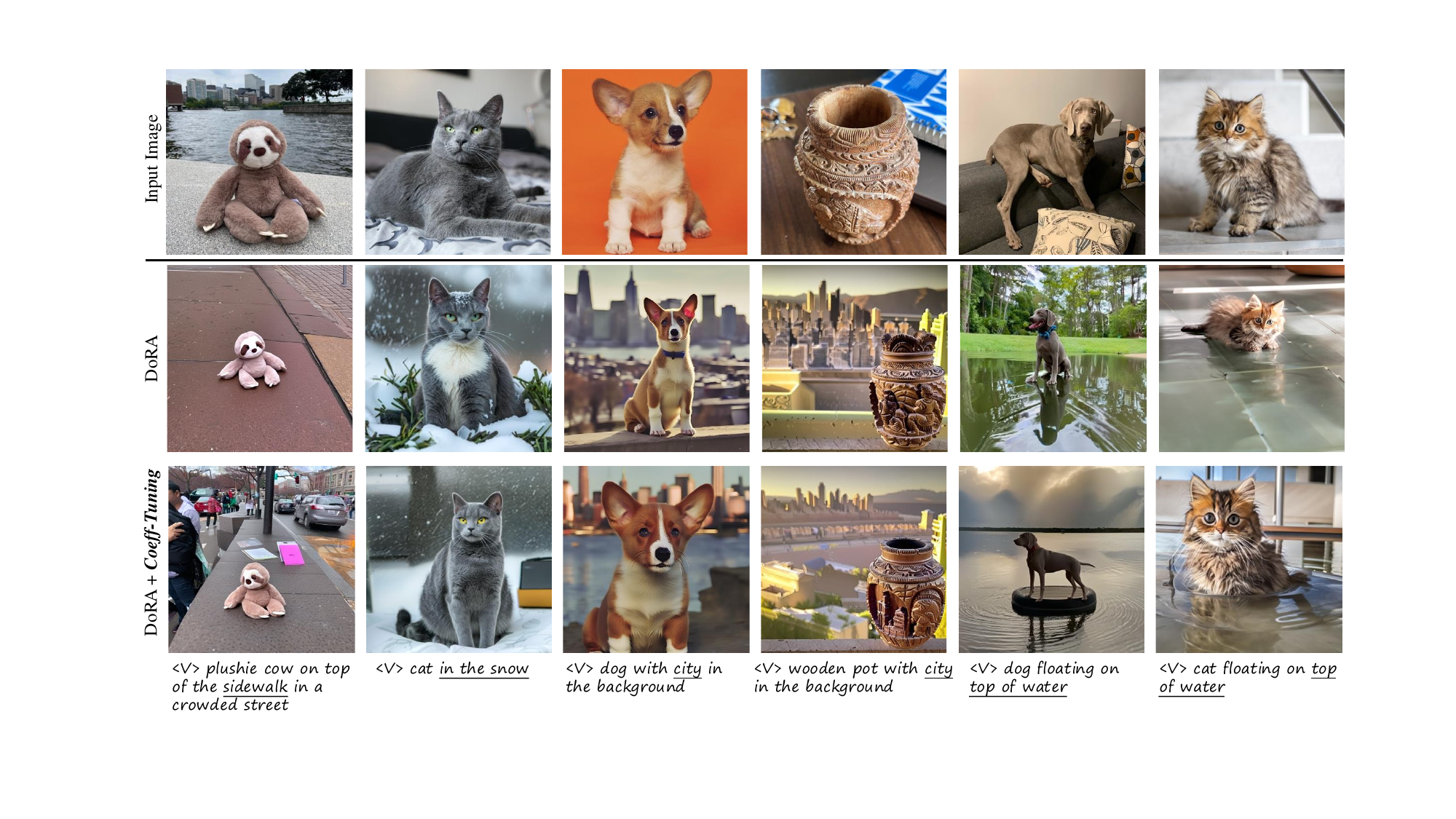}
    \caption{Results on Concept Customization.}
    \label{fig:db_4}
\end{figure*}

\begin{figure*}[t]
    \centering
    \includegraphics[trim={90pt 60pt 75pt 40pt}, clip, width=0.98\linewidth]{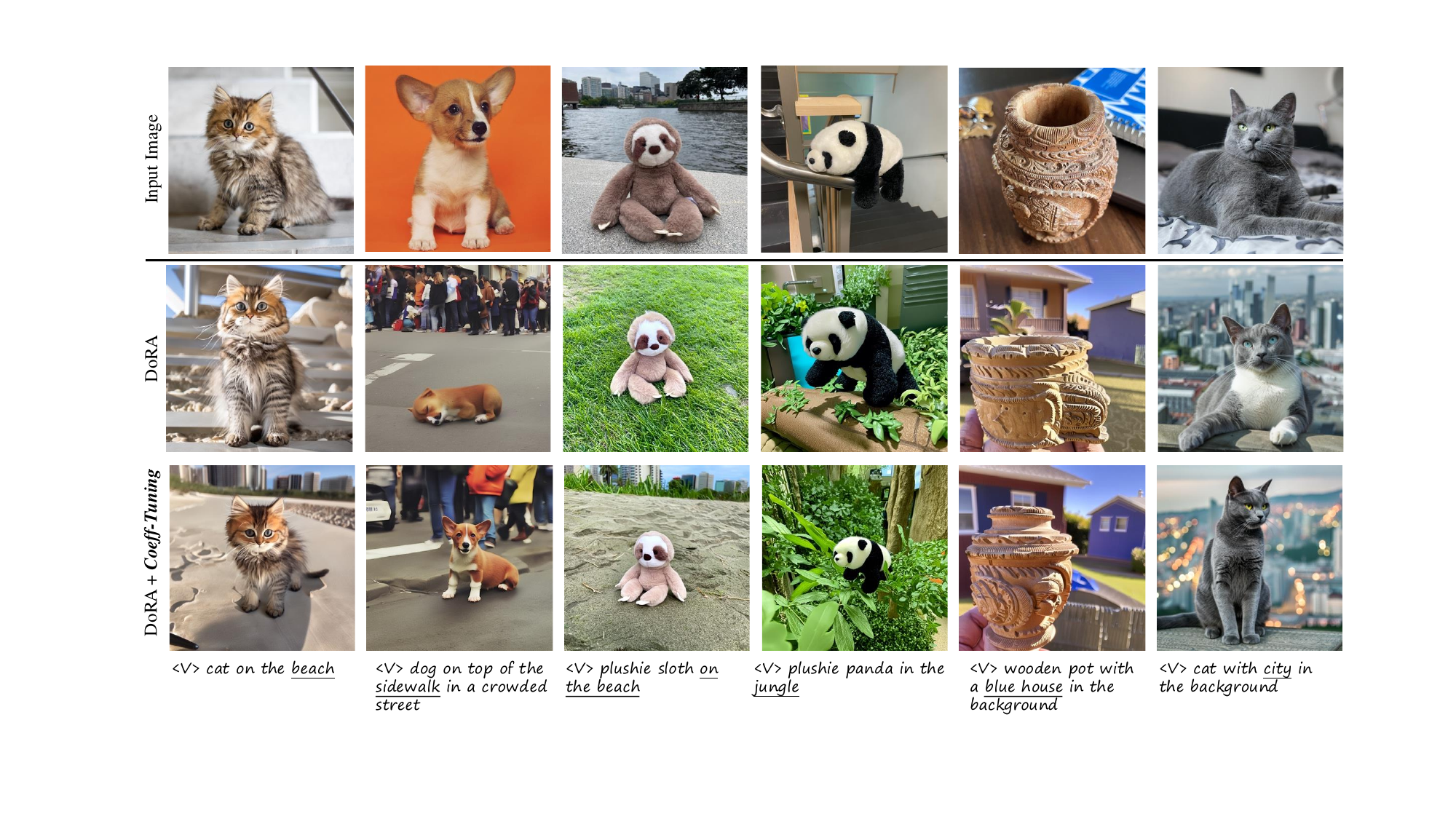}
    \caption{Results on Concept Customization.}
    \label{fig:db_5}
\end{figure*}

\begin{figure*}[t]
    \centering
    \includegraphics[trim={90pt 60pt 75pt 40pt}, clip, width=0.98\linewidth]{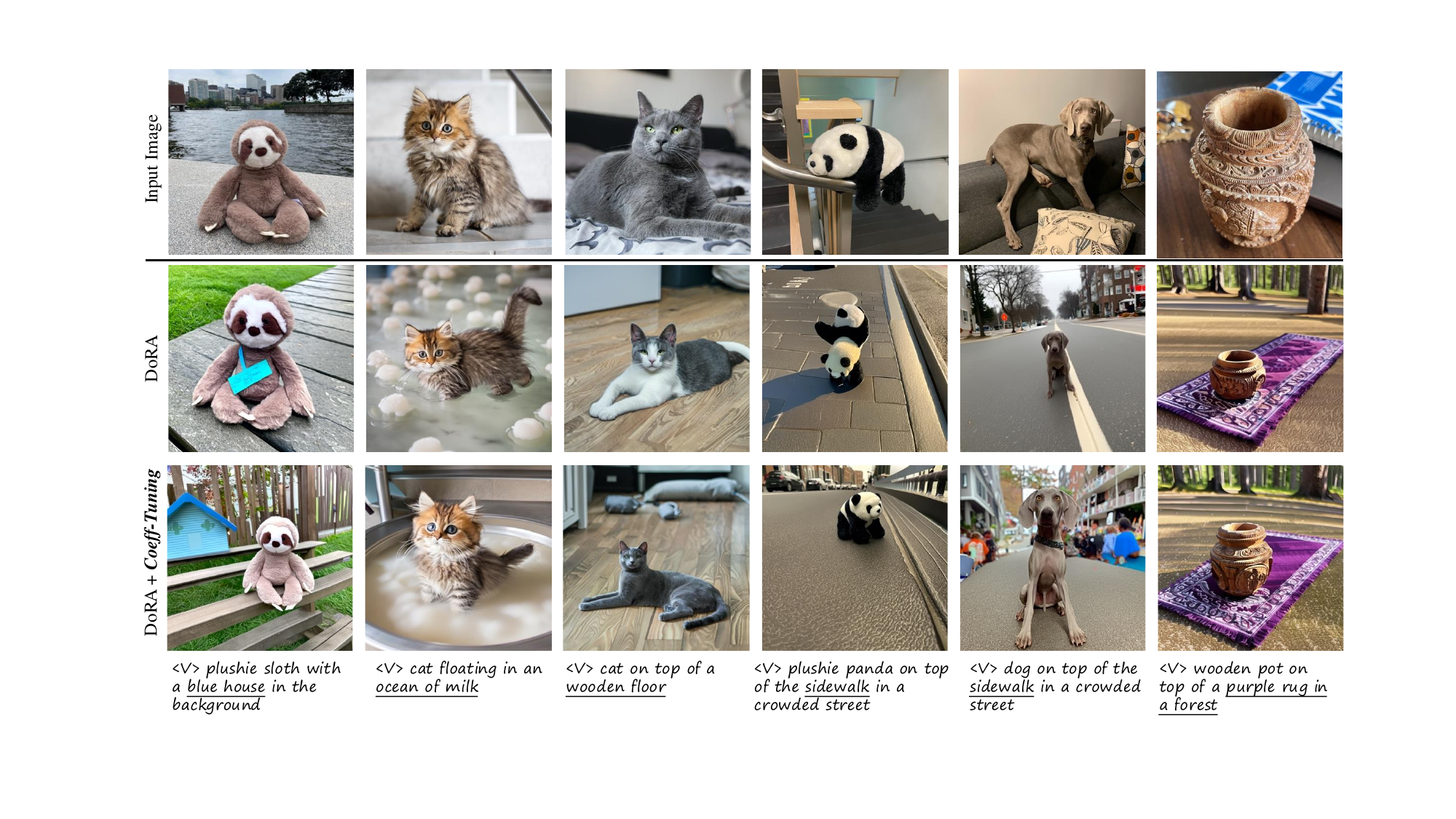}
    \caption{Results on Concept Customization.}
    \label{fig:db_6}
\end{figure*}

We provide additional comparison in Figure~\ref{fig:db_1}-\ref{fig:db_6}. The generated images with \method~have higher concept fidelity, preserving more characteristics from the input images.

\subsection{Image-Text Understanding}
\label{app:exp_image_text}
We provide the experiment details of the multi-modal tuning with VL-BART~\cite{vlbart}. Following the settings in DoRA~\cite{liu2024dora}, we utilize the fixed vision tower CLIP-ResNet-101~\cite{radford2021learning}, and tune the BART, a encoder-decoder language model, with \method~with multi-task image-text datasets as described in Sec.~\ref{sec:vlbart}. Specifically, we demonstrate the ability of the proposed \method~can be integrated with other popular weight-based PEFT methods in a plug-and-play manner. So we first add either DoRA~\cite{liu2024dora} or LoRA~\cite{hu2021lora} with $r=128$, and then add the subspace coefficient $\coeff$ in each attention layer, including self-attention layers in both the encoder and the decoder, and the cross-attention in the decoder. Specifically, we introduce additional $(12+6)*12*12=2.6K$ parameters in the BART, which are neglectable compared with the added LoRA or DoRA which takes millions of parameters. 

As for the training, we set the batch size to 300, adopt the AdamW optimizer, and train for 20 epochs. For DoRA and \method+DoRA, we adopt the learning rate of $1 \times 10^{-3}$, weight decay of $0.01$ for DoRA parameters, and learning rate $5 \times 10^{-4}$, weight decay of $1 \times 10^{-6}$, $p=0.2$ for tuning $\coeff$. For LoRA and \method+LoRA, we choose a learning rate $5 \times 10^{-4}$, weight decay of $0.01$ for LoRA parameters, and $2 \times 10^{-4}, 1\times 10^{-6}$ for coefficient $\coeff$.

\newpage

\end{document}